%% file: main.tex
\documentclass[acmtog]{acmart}

\acmSubmissionID{856}

\usepackage{multirow} % multirow table cells
\usepackage{wrapfig}
\usepackage[ruled]{algorithm2e} % For algorithms

\SetAlFnt{\small}
\SetAlCapFnt{\small}
\SetAlCapNameFnt{\small}
\SetAlCapHSkip{0pt}

\copyrightyear{2024} 
\acmYear{2024}
\setcopyright{acmlicensed}\acmConference[SA Conference Papers '24]{SIGGRAPH Asia 2024 Conference Papers}{December 3--6, 2024}{Tokyo, Japan}
\acmBooktitle{SIGGRAPH Asia 2024 Conference Papers (SA Conference Papers '24), December 3--6, 2024, Tokyo, Japan}
\acmDOI{10.1145/3680528.3687663}
\acmISBN{979-8-4007-1131-2/24/12}
\input{sections/notation}

\begin{document}

\input{sections/title.tex}

\input{sections/author.tex}

\input{sections/abstract.tex}

\input{sections/keywords.tex}

\begin{teaserfigure}
\centering
\includegraphics[width =.98\textwidth]{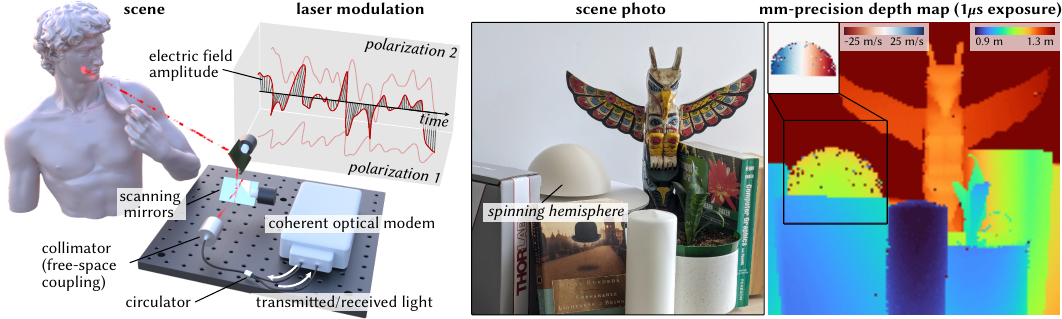}
\vspace{-6pt}
\caption{Overview of full-wavefield lidar. \textbf{(left)} Our approach repurposes an off-the-shelf coherent optical modem---typically used for telecommunications---for coherent lidar. The modem modulates the amplitude and phase of light from a 1550 nm laser in two linear polarization states. The light is emitted through a fiber optic cable, free-space collimator, and scanning mirrors, and illuminates a target. The reflected light is coupled into the fiber and directed to the receiver through a circulator. The modem, in this configuration, uses homodyne interferometry to measure the amplitude and phase of light in orthogonal polarization states. 
\textbf{(right)} Based on these measurements, we jointly estimate mm-scale 3D geometry and velocity of dynamic objects with just 1~$\mu$s per-pixel exposure time and an eye-safe transmit power of 10 mW. Depth and velocity maps are acquired with 2 mm and 0.9 m/s resolution.
}\label{fig:teaser}\vspace{-0pt}
\end{teaserfigure}

\maketitle

\input{sections/intro.tex}
\input{sections/related.tex}
\input{sections/modemimaging.tex}
\input{sections/optimization.tex}

\input{sections/results.tex}
\input{sections/discussion.tex}
 \input{sections/acknowledgments.tex}

\bibliographystyle{ACM-Reference-Format}
\bibliography{main}

\input{sections/figs.tex}

\setcounter{figure}{0}
\setcounter{section}{0}
\setcounter{equation}{0}

\input{sections/supp.tex}

\end{document}

%% file: sections/notation.tex
\colorlet{BLACK}{black}
\newcommand{\parsa}[1]{\textcolor{black}{#1}}
\newcommand{\parsarev}[1]{\textcolor{black}{#1}}
\newcommand{\parsafinal}[1]{\textcolor{black}{#1}}

\newcommand{\field}{e}
\newcommand{\inputcode}{x}
\newcommand{\outputcode}{y}
\newcommand{\freq}{\omega}
\newcommand{\timeshift}{\tau}
\newcommand{\discretetimeshift}{\Delta}

\newcommand{\freqshift}{\nu}
\newcommand{\jones}{r}

\newcommand{\jonesmatrix}{\matrixSym{\jones}}

\newcommand{\pulseshapingfilt}{b}

\newcommand{\pixelindexi}{i}
\newcommand{\pixelindexj}{j}
\newcommand{\copies}{S}
\newcommand{\copyindex}{s}

\newcommand{\numsamples}{N}
\newcommand{\samplingperiod}{T}
\newcommand{\samplingfreq}{\freq}

\newcommand{\inputcodematrix}{\matrixSym{\inputcode}}
\newcommand{\inputcodevector}[1]{\vectorSym{\inputcode}_{#1}}
\newcommand{\inputcodevectortran}{\vectorSym{\inputcode}(t)}

\newcommand{\outputcodevector}[1]{\vectorSym{\outputcode}_{#1}}

\newcommand{\txfield}{\vectorSym{\field}_\text{TX}}
\newcommand{\rxfield}{\vectorSym{\field}_\text{RX}}
\newcommand{\lofield}{\vectorSym{\field}_\text{LO}}
\newcommand{\lopower}{P_\text{LO}}

\newcommand{\txpower}{P_\text{TX}}

\newcommand{\pulseshapingfunc}{\funcsym{\pulseshapingfilt}}
\newcommand{\argmin}{\mathsf{argmin}}
\newcommand{\argmax}{\mathsf{argmax}}
\newcommand{\prob}{\mathsf{Pr}}
\newcommand{\noise}{\eta}

\newcommand{\matrixSym}[1]{\mathbf{\MakeUppercase{#1}}}
\newcommand{\vectorSym}[1]{\mathbf{\MakeUppercase{#1}}}
\newcommand{\funcsym}[1]{\mathsf{\MakeUppercase{#1}}}

%% file: sections/title.tex
\title{Coherent Optical Modems for Full-Wavefield Lidar}

%% file: sections/author.tex
%% Camera ready
\author{Parsa Mirdehghan}
\affiliation{
  \institution{University of Toronto and Vector Institute}
  \country{Canada}}
  \email{parsa@cs.toronto.edu}

\author{Brandon Buscaino}
\affiliation{
  \institution{Ciena Corporation}
  \country{USA}}
  \email{bbuscain@ciena.com}

\author{Maxx Wu}
\affiliation{
  \institution{University of Toronto and Vector Institute}
  \country{Canada}}
  \email{wumaxx@cs.toronto.edu}

\author{Doug Charlton}
\affiliation{
  \institution{Ciena Corporation}
  \country{Canada}}
  \email{dcharlto@ciena.com}
  
\author{Mohammad E. Mousa-Pasandi}
\affiliation{
  \institution{Ciena Corporation}
  \country{Canada}}
  \email{mpasandi@ciena.com}

\author{Kiriakos N. Kutulakos}
\affiliation{
  \institution{University of Toronto and Vector Institute}
  \country{Canada}}
  \email{kyros@cs.toronto.edu}

\author{David B. Lindell}
\affiliation{
  \institution{University of Toronto and Vector Institute}
  \country{Canada}}
  \email{lindell@cs.toronto.edu}

%% file: sections/abstract.tex
\begin{abstract}
The advent of the digital age has driven the development of coherent optical modems---devices that modulate the amplitude and phase of light in multiple polarization states. \enlargethispage{18pt}
These modems transmit data through fiber optic cables that are thousands of kilometers in length at data rates exceeding one terabit per second.
This remarkable technology is made possible through near-THz-rate programmable control and sensing of the full optical wavefield.
While coherent optical modems form the backbone of telecommunications networks around the world, their extraordinary capabilities also provide unique opportunities for imaging. 
Here, we repurpose off-the-shelf coherent optical modems to introduce full-wavefield lidar: a type of random modulation continuous wave lidar that simultaneously measures depth, axial velocity, and polarization. 
We demonstrate this modality by combining a 74 GHz-bandwidth coherent optical modem with free-space coupling optics and scanning mirrors.
We develop a time-resolved image formation model for this system and formulate a maximum-likelihood reconstruction algorithm to recover depth, velocity, and polarization information at each scene point from the modem's raw transmitted and received symbols. 
Compared to existing lidars, full-wavefield lidar promises improved mm-scale ranging accuracy from brief, microsecond exposure times, reliable velocimetry, and robustness to interference from ambient light or other lidar signals.\enlargethispage{6pt}
\end{abstract}

%% file: sections/keywords.tex
\begin{CCSXML}
<ccs2012>
<concept>
<concept_id>10010147.10010178.10010224</concept_id>
<concept_desc>Computing methodologies~Computer vision</concept_desc>
<concept_significance>500</concept_significance>
</concept>
<concept>
<concept_id>10010147.10010371.10010382.10010236</concept_id>
<concept_desc>Computing methodologies~Computational photography</concept_desc>
<concept_significance>500</concept_significance>
</concept>
<concept>
<concept_id>10010147.10010178.10010224.10010226.10010239</concept_id>
<concept_desc>Computing methodologies~3D imaging</concept_desc>
<concept_significance>500</concept_significance>
</concept>
</ccs2012>
\end{CCSXML}

\ccsdesc[500]{Computing methodologies~Computer vision}
\ccsdesc[500]{Computing methodologies~Computational photography}
\ccsdesc[500]{Computing methodologies~3D imaging}

\keywords{\parsa{coherent lidar, coherent optical modems, 3D imaging, Doppler imaging, polarization imaging}}

%% file: sections/intro.tex
\vspace*{-3pt}\section{Introduction}\label{sec:introduction}

Coherent optical modems are conventionally used to send digital signals over fiber optic cables by modulating the phase and amplitude of coherent light~\cite{ip2008coherent,roberts2014high,proakis2008digital}.  
Driven by the ever-increasing demands for higher networking bandwidths, these modems can now modulate and sample light at staggering rates---up to 100~GHz~\cite{roberts2017beyond}---across two orthogonal linear polarizations simultaneously~\cite{ip2007digital}. In effect, modern coherent optical modems achieve \emph{near-THz-rate, programmable control and sensing of the full optical wavefield}, with a reliability that already supports communication over optical fibers spanning thousands of kilometers.\enlargethispage{6pt}

The extreme abilities of these devices to manipulate and sense light within a fiber raise the question: how can we leverage off-the-shelf optical modems to advance the state of the art in free-space imaging?
As a first step toward addressing this question, we introduce \textit{full-wavefield lidar (FWL)}. This sensing modality falls broadly within the class of random modulation continuous wave (RMCW) lidar; however, compared to conventional RMCW techniques~\cite{sambridge2021detection} that use binary phase~\cite{bashkansky2004rf} or amplitude~\cite{ai2011high} modulation, our approach employs controllable modulation of the full coherent wavefield across two polarization states. 
Further, our FWL reconstruction algorithm provides simultaneous estimates of depth and velocity without requiring specialized encoding schemes or trading off bandwidth~\cite{banzhaf2021phase}.

To realize FWL, we use free-space coupling optics and a conventional galvanometer to turn a 400 Gb/s off-the-shelf coherent optical modem\footnote{Ciena Corporation WaveLogic 5 Nano (\url{https://www.ciena.com/products/wavelogic/wavelogic-5/nano}).} into a coherent lidar system that raster-scans the field of view (see Figure~\ref{fig:teaser}).
We develop a time-resolved image formation model that captures the unique properties of the raw output of optical modems repurposed for free-space imaging---including internal reflections, Doppler shifts, and the scrambled polarization state of back-scattered light---and use this model to formulate a maximum-likelihood reconstruction algorithm.
Our algorithm relies on the modem's raw output to solve an inverse problem that jointly recovers depth, velocity,  and polarization information.

Compared to existing lidars, FWL promises significantly more flexibility and control over the transmitted waveforms of light; mm-scale ranging; reliable velocimetry; improved performance at very short (e.g., microsecond) exposure times with eye-safe transmit power; and robustness to interference from ambient light or other lidar signals. 
We demonstrate full-wavefield lidar by capturing a variety of challenging scenes with moving objects, partial transparencies, strong ambient light, and specular surfaces.

%% file: sections/related.tex
\section{Background and Related Work}
Our work relates to multiple types of lidar and to other sensing modalities. 
We provide an overview of the connections to incoherent lidar, coherent lidar, and to optical telecommunications technologies. 

\begin{figure}
    \includegraphics[width=\columnwidth]{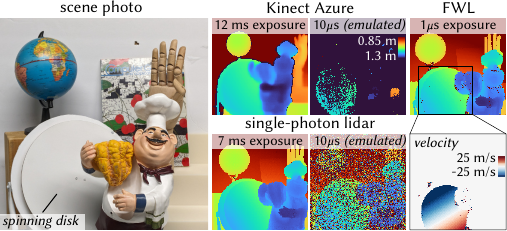}
    \vspace{-2em}
    \caption{Qualitative comparison of FWL to the Kinect Azure~\cite{bamji2018impixel} and single-photon lidar. FWL recovers accurate depth and velocity with only 1 $\mu$s  exposure times per pixel and an eye-safe 2 mW laser. The Kinect fails at light levels corresponding to 10 $\mu$s exposure times, which we emulate using neutral density filters. For the single-photon lidar system (see the supplement for a description), we emulate a 10 $\mu$s exposure time by thinning the detected photon counts~\cite{lewis1979simulation}.}\label{fig:related}  
    \vspace{-1.24em}
\end{figure}

\subsection{Incoherent Lidar}
Most commercial lidar systems operate on a principle of incoherent detection.
These lidars modulate the intensity of light to recover scene geometry by measuring phase delays of a sinusoidally modulated signal~\cite{coddington2009rapid,schuhler2006frequency} or propagation delays of emitted and backscattered laser pulses~\cite{rapp2020advances,takeuchi1983random,lin2004chaotic}. 
Incoherent lidars can also capture polarization information of backscattered light~\cite{baek2021polarimetric,baek2022all}.
However, incoherent detection schemes are sensitive to interference from other light sources or ambient light (e.g., from other lidars or the sun). 
While incoherent continuous wave systems can measure velocity from phase shifts due to the Doppler effect~\cite{hu2022differential,heide2015doppler}, pulsed systems are not sensitive to phase information and cannot be used for velocimetry in the same fashion.
FWL recovers accurate depth with 1~$\mu$s exposure times that are 10,000$\times$ shorter than that of incoherent, intensity-modulated depth sensors (e.g., the Kinect Azure~\cite{bamji2018impixel}, which uses $\approx$10 ms exposure times---see \parsa{Figure}~\ref{fig:related}). 
\parsa{Also, similar to pulsed systems~\cite{o2017reconstructing} and some continuous-wave systems~\cite{kadambi2013coded,peters2015solving}, our approach recovers time-resolved profiles of backscattered light.}\enlargethispage{6pt}

\setlength{\columnsep}{10pt}%
\begin{wrapfigure}[15]{r}{1.187 in}
    \vspace{-1.5em}
    \includegraphics[width=1.187 in]{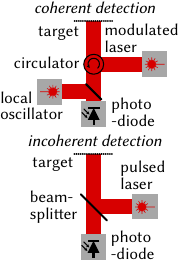}
    \vspace{-2em}
    \caption{Illustration of coherent and incoherent detection schemes.}\label{fig:detection}
\end{wrapfigure}

\paragraph{Single-photon lidar.} 
Incoherent lidars based on single-photon detection are notable for their extreme sensitivity to individual particles of light~\cite{kirmani2014first,lindell2018single}. 
However, the advantages of single-photon lidar are primarily in the weak signal regime where photons arrive infrequently at rates that are much lower than the detector dead time, which is typically on the order of tens of nanoseconds.
At higher signal levels, photon arrivals are missed, leading to difficult-to-model, non-linear distortions in photon arrival times that skew ranging estimates~\cite{rapp2021high}.

Our system functions robustly with exposures of 1~$\mu$s or less; at such short exposure times, a typical single photon lidar in the linear, low-flux regime might detect less than one laser photon on average.
Further, any received photons could be obscured by detections from ambient light.
This makes single-photon lidar very challenging when dealing with very short exposure times and ambient light. 

\vspace*{-3pt}
\subsection{Coherent Lidar}
Coherent lidar detects the amplitude and/or phase of backscattered incident laser light by interfering it with unmodulated light from the same laser (referred to as the local oscillator~\cite{henderson2005wind}), or from another laser at a different frequency~\cite{li2021exploiting}.
In contrast to incoherent lidar or other techniques such as optical coherence tomography~\cite{gkioulekas2015micron,kotwal2023passive}, it is critical that the laser source have a high degree of temporal coherence so that the incident light and local oscillator remain correlated when they are interfered at a photodiode, as illustrated in Figure~\ref{fig:detection}. 
Please refer to Goodman~\shortcite{goodman2015statistical} for a detailed treatment of temporal coherence; we provide a mathematical description of coherent detection in the supplement.\enlargethispage{6pt}   

In general, coherent lidar systems have several advantages compared to their incoherent counterparts. 
Since they use continuous wave emission, they can allow eye-safe operation at higher average optical powers compared to pulsed lasers, which may have very high peak power depending on the duty cycle.
Additionally, their use of coherent averaging (i.e., of the complex electric field) results in a linear increase in signal-to-noise ratio (SNR) with exposure time compared to the square-root relation of incoherent averaging of intensity measurements~\cite{baumann2019signal}. 
Further, coherent detection strongly suppresses interference from ambient light due to the interferometric detection procedure and use of balanced photodetectors~\cite{behroozpour2017lidar}. \enlargethispage{6pt}

\paragraph{FMCW lidar.} 
Perhaps the most common type of coherent lidar is based on a frequency-modulated continuous wave (FMCW) transmit signal~\cite{zheng2005optical}.
Specifically, FMCW lidars transmit a ``chirp'' signal whose optical frequency increases linearly with time. 
\parsa{While one advantage of FMCW lidar is the simplicity of its modulation scheme (i.e., the linear frequency sweep), the system performance is strongly coupled to the modulation waveform. FMCW depth resolution is tied to the bandwidth of the linear frequency sweep, and the maximum depth and velocity resolution depend on the duration of the sweep~\cite{behroozpour2017lidar}. For many swept-source lasers, there is limited flexibility to adjust these parameters (e.g., beyond certain preset configurations). In contrast, our system’s performance is easily configurable and largely decoupled from the modulation waveform: the depth resolution is tied to the sample rate (74 GHz for our system), and the maximum depth and the velocity resolution depend on the exposure time—each of which can be continuously controlled. Our setup is also more general as it measures the response of a scene to arbitrary, programmable wavefields.}
\parsa{Additionally}, FMCW lidars are sensitive to interference from other frequency-modulated lidars\parsa{~\cite{hwang2020mutualfmcw}}. \parsa{RMCW techniques, including our proposed FWL system, are less susceptible to interference than FMCW because the transmit waveforms are likely to be orthogonal to any interfering signal~\cite{chen2023breaking, hwang2020mutual}.}

\paragraph{RMCW lidar.} 
\parsa{FWL can be considered as a type of RMCW lidar~\cite{takeuchi1983random}; however, different from conventional RMCW lidar, we do not use a binary modulation of phase~\cite{bashkansky2004rf,sambridge2021detection} or amplitude~\cite{ai2011high,takeuchi1986diode}. 
Instead, we use discrete, pseudorandom \nobreak modulation of both amplitude and phase in two polarization states.
In this sense, our work is close to ``true random'' or ``chaos'' lidars that use amplified spontaneous emission noise~\cite{hwang2020mutual} or chaotic microcombs~\cite{chen2023breaking} to induce random fluctuations in the amplitude of emitted light. 
However, unlike FWL, their output waveforms cannot be programmably controlled since they are governed by stochastic phenomena.
} 

\parsa{Many of the drawbacks associated with FMCW lidar (e.g., related to configurability or interference as discussed above) can be mitigated with RMCW lidar---though there are considerable implementation challenges.}
The range resolution is determined, in part, by the modulation speed and sample rate---typically tens of GHz to achieve mm-scale resolution. 
Thus, few examples of this modality exist in the literature due to the significant hardware requirements related to ultrafast sample rates and the computational challenge of modeling Doppler shifts, laser phase noise, speckle, and polarization changes induced by scattering.

Our work overcomes hardware challenges associated with RMCW lidar by leveraging existing, off-the-shelf optical modems used for telecommunications.
We demonstrate that FWL improves the accuracy of ranging and velocimetry over other modulation schemes (e.g., without phase modulation or amplitude modulation) implemented on the same optical modem. 
We also show that our polarization-aware reconstruction framework improves accuracy and performance at low SNRs compared to matched-filtering schemes similar to those used in RMCW lidar.

\begin{figure*}[t]
\includegraphics[width=\textwidth]{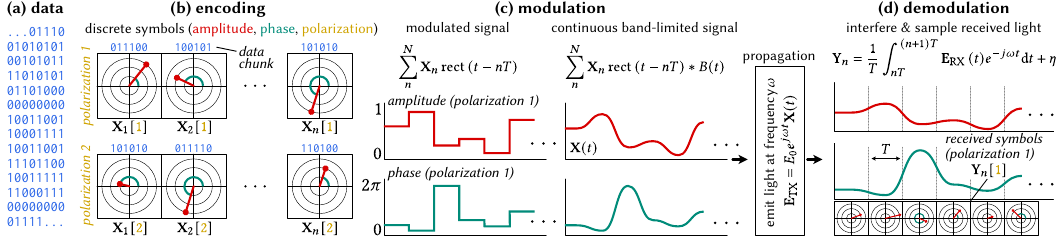}
\vspace*{-2em}
\caption{Overview of data transmission in coherent optical modems. \textbf{(a,b)} Binary data are collected and encoded into a discrete sequence of symbols $\inputcodevector{n}$, each paired with a certain amplitude, phase, and polarization state of light. 
\textbf{(c)} This sequence of symbols is used to create a piecewise constant waveform with segments of duration $T$ (shown for a single polarization).
In practice, a band-limited version of this waveform modulates laser light with amplitude $E_0$ and wavelength $\lambda = c \frac{2\pi}{\samplingfreq}$, where $\samplingfreq / 2\pi$ is the optical frequency and $c$ is the speed of light. 
\textbf{(d)} The modulated light is emitted, collected by a receiver, and interfered with an unmodulated receiver-side laser to remove the optical frequency shift. The resulting waveform is sampled to recover the demodulated symbols $\mathbf{Y}_{n}$.}\label{fig:modem}
\vspace{-1em}
\end{figure*}

\vspace*{-2pt}\enlargethispage{6pt}
\subsection{Optical Telecommunications}
Our work makes use of coherent optical modems that are conventionally used to send digital signals over fiber optic cables~\cite{ip2008coherent,roberts2014high}. 
Typically, these modems use a modulation scheme~\cite{proakis2008digital} to optically encode digital information in the amplitude, phase, polarization, and frequency of transmitted light. 
Signals from optical modems are often combined with wavelength division multiplexing~\cite{brackett1990dense} and transmitted in parallel across a single-mode fiber at different wavelengths (e.g., 1530--1565 nm).  
Polarization division multiplexing is also used to transmit two signals in parallel using orthogonal linear polarizations of the electric field~\cite{ip2007digital}.

We operate a coherent optical modem on a single wavelength channel at 1550 nm; this wavelength has the benefit of being eye safe at roughly 50$\times$ higher transmit powers (up to 10 mW) compared to visible wavelengths because light is absorbed at the cornea rather than propagating to the retina~\cite{sliney2013safety}.  

\sloppy{We use a simple modulation scheme consisting of random amplitude and phase values sampled from a complex Gaussian distribution---the optimal scheme for measurements corrupted by Gaussian noise~\cite{forney1992trellis}.
Modulated light is transmitted on two orthogonal polarization channels which are coupled to free space and backscattered from surfaces with different material properties.
As such, the transmitted light signals are distorted by speckle, and polarization and phase information is scrambled. 
We explicitly estimate these distortions to recover depth and velocity.}

%% file: sections/modemimaging.tex
\section{Coherent Optical Modem Imaging}

Coherent modems are designed to enable high-bandwidth transmission of data over optical fiber. 
Achieving data rates of many gigabits per second necessitates exploiting multiple degrees of freedom in the transmitted light. 
In this section, we outline the working principle of optical modems---from encoding digital information into discrete symbols, to transmitting, receiving, and demodulating the digital data. 
Then, we outline how optical modems can be repurposed for imaging in free space.

\subsection{Coherent Modulation and Demodulation}\enlargethispage{6pt}

A coherent optical modem realizes two main functionalities: \emph{modulation} of a coherent laser, where a predefined data sequence is encoded into a laser's electric field; and \emph{demodulation} of received light, where the transmitted data sequence is inferred from a measured electric field (see Figure~\ref{fig:modem}).

\paragraph{Modulation.} 
We consider a coherent modem that modulates the phase and amplitude of light in two orthogonal linear polarization states. 
Given an input sequence of digital data (Figure~\ref{fig:modem}), we can define a complex-valued matrix $\inputcodematrix \in \mathbb{C}^{2 \times \numsamples}$ that is used to map digital data to different output states of the coherent laser. 
More specifically, each column of this matrix, denoted by $\inputcodevector{n}$, contains two complex-valued \emph{symbols} that specify the phase and amplitude of the outgoing electric field at each polarization during the $n$th output time interval, or \textit{symbol period}: $n\samplingperiod\leq t< (n+1)\samplingperiod$. 

Once the digital data are encoded into symbols, the modulation process involves two steps. 
First, the discrete sequence of symbols $\inputcodevector{n}$ is transformed into a continuous coded waveform $\inputcodevectortran$:
\begin{equation}
	\vspace*{-2pt}
	\inputcodevectortran = \sum_{n=0}^{\numsamples-1} \inputcodevector{n} \: \mathsf{rect} \bigg( t - n\samplingperiod \bigg) * \pulseshapingfunc (t),
\end{equation}

\noindent where $\mathsf{rect} (t)$ is a rectangular function that is equal to one for $0\leq t < \samplingperiod$ and zero elsewhere, and $\pulseshapingfunc (t)$ is a filter that creates a smooth, band-limited signal from the piecewise concatenation of rectangular functions. 
Second, the coded waveform modulates a laser of wavelength $\lambda=c\frac{2\pi}{\samplingfreq}$ whose electric field propagates at the speed of light $c$ and oscillates with carrier frequency $\samplingfreq/ 2\pi$ and amplitude $\mathrm{E}_0$. The transmitter output of the coherent optical modem can then be formulated as
\begin{equation}
	\txfield(t) = \mathrm{E}_0 e^{j \samplingfreq t} \, \mathbf{X}(t),
\end{equation}
such that the digital information is completely encoded in the transmitted electric field $\txfield$.

\paragraph{Demodulation.} The goal of demodulation is to recover an estimate $\outputcodevector{n}$ of the transmitted symbol sequence $\inputcodevector{n}$ by measuring the amplitude and phase of received electric field $\rxfield (t)$. Either \emph{homodyne} or \emph{heterodyne} coherent detection can be used, where the measured electric field is interfered with a laser source called the \textit{local oscillator}. In the context of optical communications, the transmitted signal and the local oscillator are generated by different laser sources; the phase of the local oscillator is matched to that of the received signal using a phase-locked loop or feed-forward carrier synchronization, which maintains temporal coherence~\cite{ip2008coherent}.

In a homodyne detection scheme, the received signal and local oscillator have the same carrier frequency $\samplingfreq/2\pi$.
Interference of these two sources \textit{downconverts} the received signal: the high-frequency oscillations of the electric field at the laser frequency are removed, and the modulated waveform containing the encoded digital data is recovered (refer to the supplement for a mathematical description of this procedure). 
The detected sequence $\mathbf{Y}_{n}$ is given as:\enlargethispage{6pt}
\begin{equation}
	\mathbf{Y}_{n} = \frac{1}{\samplingperiod} \int_{n\samplingperiod}^{(n+1)\samplingperiod} \rxfield (t) \; e^{-j\samplingfreq t} dt + \noise.\label{eq:detection}
\end{equation}
Here, the received electric field after interference with the local oscillator is sampled via integration over the symbol period. 
The term $\noise$ is complex Gaussian noise that incorporates multiple effects, including thermal noise in the receiver, shot noise, and noise due to inline optical amplifiers~\cite{ip2008coherent}. 

We note that this section provides a simplified description of the coherent modulation and demodulation procedure. 
In practice, optical modems contain additional optical and digital processing stages to ensure that the local oscillator is locked to the transmit laser and that the received signal is sampled with the correct timing. 
\parsarev{While shot noise generally follows a Poisson distribution, here it is approximated by a Gaussian distribution.}
Further, we note that the Gaussian noise model neglects secondary effects such as optical fiber non-linearities~\cite{singh2007nonlinear}.
For a detailed treatment of coherent detection in optical modems, please refer to Ip et al.~\shortcite{ip2008coherent}.

\begin{figure*}[ht]
    \centering
    \includegraphics[width=0.99\textwidth]{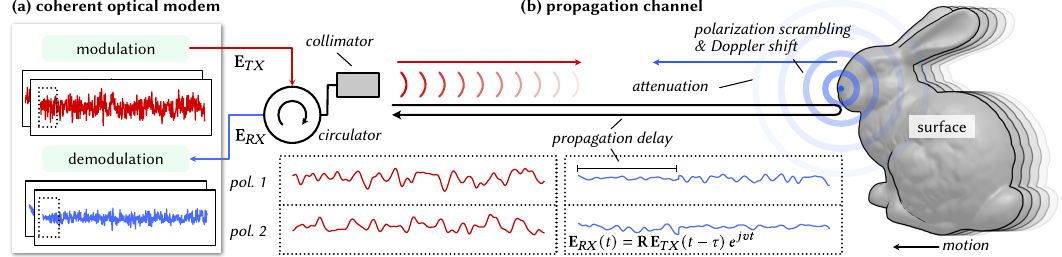}
        \vspace*{-1em}
\caption{Illustration of imaging with coherent optical modems. \textbf{(a)} The modulated wavefield $\txfield$ is transmitted to the target through a fiber optic cable, circulator, and collimator. The received wavefield $\rxfield$ is demodulated and detected by the optical modem. \textbf{(b)} The transmitted wavefield is distorted by multiple effects: the propagation delay induces a shift in the measurements, shown in the bottom plots of $\rxfield$; scattering off of a moving surface scrambles the two transmitted polarization channels (modeled by multiplication with a Jones matrix $\jonesmatrix$) and induces a Doppler shift of frequency $\freqshift$; the wavefield is attenuated as it propagates back to the collimator; last, the measurements are corrupted by noise $\noise$ from the optical modem or optical amplifiers (not shown).}\label{fig:freespace}
\vspace{-1em}
\end{figure*}

\subsection{Repurposing Coherent Optical Modems for Imaging}
Our goal is to repurpose coherent optical modems for the task of free-space 3D imaging. 
Given a known transmit symbol sequence and the received symbol sequence, we aim to infer unknown scene parameters, namely depth, radial velocity, and change in polarization state. 
In contrast to communication systems where the transmitted sequence is unknown and the goal is to recover the digital data, we seek to understand how transmitted waveforms are distorted by the propagation channel and, in turn, recover scene properties. 

\paragraph{System overview.} Figure~\ref{fig:freespace} provides an overview of the proposed system. The optical modem modulates the laser electric field based on an arbitrary, known symbol sequence, and the light is transmitted to the scene through a fiber optic cable and collimator. 
The reflected light is collected by the same collimator and directed to the optical modem. 
In our setup, the transmitter and receiver share the same light propagation path through a circulator (which separates optical signals traveling in opposite directions), and we use a set of scanning mirrors to image the scene (Figure~\ref{fig:teaser}).\enlargethispage{6pt}

\paragraph{Measurement model.} 
The received and demodulated electric field $\rxfield(t)$ is \parsarev{altered} by three main effects resulting from propagation to the scene and back. 
First, the demodulated electric field is time-shifted relative to the transmit signal due to the propagation delay to the target and back.
Second, if illuminating a moving target, the field is frequency-shifted due to the Doppler effect~\cite{mcmanamon2015field}. 
Finally, the field is distorted by attenuation and changes in the polarization state due to the surface reflection. 
The received electric field can therefore be modeled as 
\begin{equation}
	\rxfield(t) = \jonesmatrix \; \txfield(t-\timeshift) \; e^{j\freqshift t}.
	\label{eq:formation}
\end{equation}
The terms $\timeshift$ and $\freqshift$ denote the propagation delay and frequency shifts of the demodulated electric field. 
The Jones matrix $\jonesmatrix$ is a $2\times2$ complex-valued matrix that describes how the transmitted electric field is transformed by a polarization-dependent attenuation and rotation induced by the properties of the target surface and propagation through the fiber~\cite{gordon2000pmd}). 
\parsa{We will later show that the optical modem's measurement of two polarization channels enables recovery of the Jones matrix.} 

In practice, we assume that the received, downconverted electric field is approximately constant over the symbol integration period, allowing us to drop the integral in Equation~\ref{eq:detection} and relate the received symbol sequence $\outputcodevector{n}$ directly to the transmitted sequence as 
\begin{equation}
    \outputcodevector{n} \; = \; \jonesmatrix \; \inputcodevector{n-\discretetimeshift} \; e^{j\freqshift t} + \noise,
	\label{eq:detection2}
\end{equation}
where $\discretetimeshift = \lfloor \timeshift / \samplingperiod\rfloor$ is the integer shift in the symbol sequence due to the propagation delay. 
This integer approximation to the time delay (i.e., $\timeshift/\samplingperiod \approx \lfloor \timeshift / \samplingperiod\rfloor$) is justified in our case, where $\timeshift \gg \samplingperiod$.\enlargethispage{6pt}

\begin{figure}[t]
    \includegraphics[width=\columnwidth]{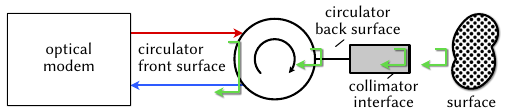}
     \vspace*{-2em}
    \caption{Our FWL prototype and coaxial lidar systems in general have non-idealities such as reflections from interfaces in the optical path.}\label{fig:reflections}
    \vspace{-1em}
\end{figure}

Coaxial lidar configurations like ours exhibit inter-reflections caused by the various interfaces in the optical propagation path.
Additional reflections from partially transparent surfaces in the scene are also possible.
Thus measurements from our system (Figure~\ref{fig:freespace}) can be modeled as a superposition of signals, including multiple inter-reflections and scene reflections as depicted in Figure~\ref{fig:reflections}. 
While we are mainly interested in the reflection off of target surfaces, other reflections are captured, e.g., from the circulator interfaces, and from the glass--air interface of the collimator. 
Since these reflections due to internal surfaces are static, they do not induce a Doppler shift. 
We therefore consider a generalization of Equation~\ref{eq:detection2} that models the received symbol sequence as a discrete superposition of the delayed, polarization-scrambled, and potentially frequency-shifted copies of the transmitted signal:
\begin{equation}
    \outputcodevector{n} \; =  \;  \sum_{\copyindex=0}^{\copies-1} \jonesmatrix_{\copyindex} \; \inputcodevector{n-\discretetimeshift_\copyindex} \; e^{j\freqshift_\copyindex t}\;  + \; \noise.
	\label{eq:detection3}
\end{equation}
Here, $\copyindex$ denotes the index of each copy of the transmitted signal; $\jonesmatrix_{\copyindex}$ and $\discretetimeshift_\copyindex = \lfloor \timeshift_\copyindex / \samplingperiod\rfloor$ model the corresponding polarization scrambling, signal attenuation, and propagation delay; and $\freqshift_\copyindex$ is the corresponding Doppler shift (zero for internal reflections). % and $\noise$ is complex Gaussian noise.

\paragraph{Key objective.} Given a known transmitted symbol sequence, our goal is to recover per-pixel scene unknowns from the received symbol sequence through maximum likelihood estimation:
\begin{equation}
    \{\jonesmatrix^*_\copyindex,\discretetimeshift^*_\copyindex,\freqshift^*_\copyindex  \} = \underset{\{\jonesmatrix_\copyindex,\discretetimeshift_\copyindex,\freqshift_\copyindex \}}{\argmax} \: \prob \big( \outputcodevector{n} | \inputcodevector{n} \big).
	\label{eq:ml}
\end{equation}
For each pixel, $\discretetimeshift_\copyindex^*$ is an estimate of the depth $d_\copyindex = \discretetimeshift^*_\copyindex\cdot\samplingperiod c / 2$, where $c$ is the speed of light.
The velocity is calculated from the Doppler shift as $\frac{\freqshift^*_{\copyindex}\cdot c}{\freq}$, and $\jonesmatrix$ describes the attenuation and polarization change, which depend on the surface and material.

%% file: sections/optimization.tex
\section{Joint Estimation of Depth, Velocity, and Polarization}

We provide an optimization algorithm for joint estimation of depth, radial velocity, and polarization change at each pixel.

\paragraph{Matched filtering.}
In the case of a single direct reflection ($\copies=1$ in Equation~\ref{eq:detection3}), no polarization scrambling, and additive Gaussian noise, the matched filter is the maximum-likelihood estimator for recovering the unknown propagation delay of a known transmit waveform~\cite{turin1960introduction}.  
The matched filter is also the typical approach for depth estimation in RMCW lidar~\cite{takeuchi1983random,sambridge2021detection}. 
Using the notation of Equation~\ref{eq:ml}, matched filtering can be expressed as the optimization
\begin{align}
    \discretetimeshift^* = \underset{\discretetimeshift}{\argmax} \;\left| \;\sum\limits_{\discretetimeshift=-\infty}^{\infty} \overline{\inputcodematrix}_{n - \discretetimeshift} \outputcodevector{n}\;\right|^2,\label{eq:matchedfilter}
\end{align}
where $\overline{\inputcodematrix}_{n}$ is the complex conjugate of the transmitted symbol $\inputcodematrix_{n}$.\enlargethispage{6pt}

\paragraph{Joint estimation.}
While matched filtering offers a straightforward and computationally efficient solution to depth estimation, it is not well-suited for complex scenes.
Specifically, in the general case captured by Equation~\ref{eq:detection3}, no single time delay can explain the received symbol sequence because of polarization scrambling, Doppler shift, and multiple reflections.  
Crucially, by ignoring these effects, the optimization in Equation~\ref{eq:matchedfilter} provides no information about the velocity and polarization properties of scene points.
Proper handling of these effects is essential, not only for FWL, but for any lidar system---multiple reflections can be caused by depth discontinuities or partially transparent surfaces. 
In coherent lidar, Doppler shifts must be modeled for accurate depth estimation and velocimetry for dynamic scenes. 

Instead, we seek the values of $\jonesmatrix$, $\discretetimeshift$, and $\freqshift$, as well as the number of shifted copies of the transmitted symbols $\copies$, that minimize the mean squared error between the transmitted and received symbols, described as follows:
\begin{align}
    \underset{\{ \copies, \jonesmatrix_\copyindex,\discretetimeshift_\copyindex, \freqshift_\copyindex \}}{\argmin} \sum_{n=0}^{N-1} \ \left\Vert\outputcodevector{n} \; - \;\sum_{\copyindex=0}^{\copies-1} \jonesmatrix_{\copyindex} \; \inputcodevector{n-\discretetimeshift_\copyindex} \; e^{j\freqshift_\copyindex t} \right\Vert^{2}_2.
	\label{eq:opt}
\end{align}
Solving this problem provides the maximum likelihood estimate of these parameters under a Gaussian noise model. 
This approach is analogous to channel estimation employed in the digital communications literature~\cite{proakis2008digital}. 

The optimization in Equation~\ref{eq:opt} does not have a closed-form solution; the associated objective function is combinatorial in nature because of the (typically small and unknown) number of additive terms.  
To make optimization tractable, we relax the objective by discretizing the space of Doppler shifts and associating an unknown Jones matrix $\jonesmatrix_{\discretetimeshift, \freqshift}$ to each possible time delay $\discretetimeshift$  and Doppler shift $\freqshift$.

\begin{align}
    \mathcal{L}_{\text{data}} = \; \sum_{n=0}^{\numsamples-1} \left\Vert \; \outputcodevector{n} - \sum_{\discretetimeshift, \freqshift} \jonesmatrix_{\discretetimeshift,\freqshift}\,\inputcodevector{n-\discretetimeshift} \; e^{j\freqshift t}\; \right\Vert^2_2.\label{opt2}
\end{align}

Since we expect a relatively small number of contributions to the sum of Equation~\ref{eq:opt}, we introduce a sparsity-promoting \nobreak regularization term $\mathcal{L}_\text{sparse}$ into our objective function.
We also introduce a total variation spatial regularization term $\mathcal{L}_\text{TV}$ to help mitigate errors across pixels due to speckle noise~\cite{goodman2007speckle} by encouraging spatial smoothness in the energy of the reconstructed Jones matrices:
\begin{align}
    \mathcal{L}_\text{sparse} &= \sum_{\discretetimeshift, \freqshift}  \|\jonesmatrix_{\discretetimeshift,\freqshift}\|_{\text{F}},\,\, \text{and}\\
    \mathcal{L}_\text{TV} &= \sum_{\pixelindexi,\pixelindexj, \discretetimeshift, \freqshift} \sqrt{\,\lvert\left(\mathbf{D}_\text{v}\|\jonesmatrix_{\discretetimeshift,\freqshift}\|_{\text{F}}\right)_{i,j}\rvert^2 + \lvert\left(\mathbf{D}_{\text{h}}\|\jonesmatrix_{\discretetimeshift,\freqshift}\|_\text{F}\right)_{i,j}\rvert^2}.
\end{align}
Here, $\lVert\cdot\rVert_{\text{F}}$ is the Frobenius norm, $\mathbf{D}_\text{v}$ and $\mathbf{D}_\text{h}$ compute vertical and horizontal finite differences between pixels, and $i$ and $j$ index the vertical and horizontal pixel locations. 
The resulting optimization problem is
\begin{align}
    \underset{\jonesmatrix_{\discretetimeshift, \freqshift}}{\argmin} \; \mathcal{L}_\text{data} + \lambda_\text{sparse} \mathcal{L}_\text{sparse} + \lambda_\text{TV} \mathcal{L}_\text{TV},\label{eq:objective}
\end{align}
where the scalars $\lambda_\text{sparse}$ and $\lambda_\text{TV}$ weigh each regularizer.

Once the Jones matrices have been estimated for all $\discretetimeshift$ and $\freqshift$ , we solve for depth and velocity as follows. 
Given that the scene contains a single reflection from a target surface, we return the delay $\discretetimeshift$  and frequency shift $\freqshift$ whose associated Jones matrix has the maximum Frobenius norm (and we ignore delays due to internal reflections).
That is, we find 
\begin{align}
    (\discretetimeshift^{\ast},\ \freqshift^{\ast}) &= \underset{\discretetimeshift,\ \freqshift}{\argmax} \|\jonesmatrix_{\discretetimeshift,\freqshift}\|_{F} \quad 
    \mathrm{subject\: to} \quad \discretetimeshift > \discretetimeshift_\text{min},
	\label{eq:obj}
\end{align}
where $\discretetimeshift_\text{min}$ is the minimum delay due to free-space propagation, ignoring internal reflections in the optical modem\parsarev{.}\footnote{We assume that the values $\discretetimeshift$ associated with internal reflections are calibrated a priori.}\enlargethispage{6pt}

\begin{figure}[t]
\includegraphics[width=\columnwidth]{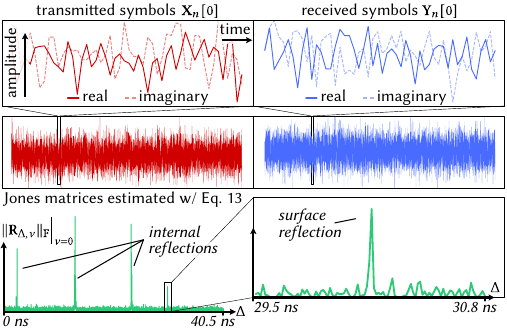}
\vspace{-2em}
\caption{Estimated Jones matrices. The transmitted and received symbols (top and middle) are related by a convolution with a sequence of Jones matrices $\jonesmatrix_{\discretetimeshift, \freqshift}$ (bottom) estimated using Equation~\ref{eq:objective}. The real and imaginary symbol components (top) describe the amplitude and phase (Fig.~\ref{fig:modem}). We plot $\lVert\jonesmatrix_{\discretetimeshift, \freqshift} \rVert_F$ (bottom) for a static scene with $\freqshift=0$.}\label{fig:optimization}
\vspace{-1em}
\end{figure}

\subsection{Implementation Details}
\paragraph{Optimization.} 
We implement the optimization in Equation~\ref{eq:objective} in PyTorch~\cite{paszke2019pytorch} and use Adam~\cite{kingma2015adam} with a learning rate of $1 \times 10^{-2}$. For the weighting of loss terms we use $\lambda_\text{sparse} = 0.1$ for static scenes and $\lambda_\text{sparse} = 0.3$ for dynamic scenes, with $\lambda_\text{TV} = 0.1$. 
In practice, to ease computational requirements, we only apply the TV loss to Jones matrices across the $\discretetimeshift$ dimension where $\nu=0$ (i.e., to the static scene components).
Similarly, for dynamic scenes we set a maximum axial velocity of 30 meters/second, and only optimize for the feasible frequency shifts. 
Please see the supplement for additional implementation details related to the optimization.\enlargethispage{2pt}

\begin{figure}
    \includegraphics[width=\columnwidth]{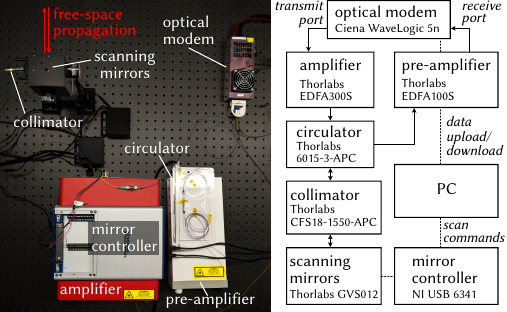}
    \vspace{-1em}
    \caption{Prototype coherent lidar system. \textbf{(left)} Each component is shown in a photo of the system. \textbf{(right)} The connections between each component are illustrated. The transmit and receive ports of the optical modem are coupled into single-mode optical fiber. The transmit cable connects through an amplifier and collimator to a set of scanning mirrors, whose positions are set by a controller. The collimator couples the reflected light through the circulator, to the pre-amplifier, and into the receive port of the optical modem. A PC interfaces with the modem and the scanning mirrors.}\label{fig:hardware}
    \vspace{-1.24em}
\end{figure}

\enlargethispage{6pt}
\paragraph{Hardware prototype.}
A photo and illustration of the hardware prototype are shown in Figure~\ref{fig:hardware}.
We use a Ciena WaveLogic 5n modem with a sampling frequency of $1/\samplingperiod = 74$ GHz (i.e., an optical path resolution of $4$ mm or depth resolution of $2$ mm); the laser operates at a wavelength of 1550 nm. 
We communicate with the modem over its QSFP-DD electrical interface~\cite{qsfp} to program the transmit sequence and read out the measured data.
The transmitted sequence length, limited by finite modem memory, is set to $\approx 2^{16}$ symbols, providing a maximum unambiguous range of approximately 130 meters.
We use an Erbium-doped fiber amplifier (EDFA~\cite{becker1999erbium}) to boost the power of the emitted laser light from 1 mW up to a maximum of 100 mW; all experiments use an eye-safe transmit laser power of 2 mW unless otherwise specified.
Another EDFA is used as a pre-amplifier to boost the power of the received light up to the level expected by the modem ($\approx$1 mW).

%% file: sections/results.tex
\section{Experimental Results}
\label{results}

We evaluate FWL across different exposure settings, and we compare to alternative modulation schemes and reconstruction techniques. 
We also demonstrate the approach for recovery of depth and velocity information, imaging through translucent media, imaging under strong ambient light, reconstructing objects with sub-surface scattering, and reconstruction of room-scale scenes. 

\vspace*{-3pt}
\subsection{Quantitative and Comparative Evaluation}

\paragraph{Generalized matched filtering.}
We compare our joint estimation framework to a straightforward generalization of matched \nobreak filtering which incorporates the multiple polarization channels of FWL.
Specifically, we modify Equation~\ref{eq:matchedfilter} to correlate the transmit and received symbols sequences across both polarization channels: 
\begin{align}
\discretetimeshift^* = \underset{\discretetimeshift}{\argmax} \;\sum_{p,q} \left| \;\sum\limits_{\discretetimeshift=-\infty}^{\infty}\overline{\inputcodematrix}_{n - \discretetimeshift}[p] \outputcodevector{n}[q]\;\right|^2, 
\end{align}
where $p, q \in \{1, 2\}$ index the polarization channels.
While this procedure is convenient because it incorporates information across polarization channels, it does not recover Doppler frequency shift, nor does it recover the complex-valued Jones matrices corresponding to reflections.\enlargethispage{6pt}

\paragraph{Comparison to other modulation schemes.} 
We compare FWL to other modulation schemes emulated with the coherent optical modem.
Specifically, we compare FWL to dual-polarization phase-only modulation, dual-polarization amplitude-only modulation, and single-polarization phase and amplitude modulation (see supplement for implementation details). 
The results are shown in Figure~\ref{fig:comparisons}; FWL recovers depth maps with fewer outliers compared to the other modalities and compared to depth estimation using generalized matched filtering. 
We assess depth precision by imaging a planar target at distances of roughly 0.5, 1.0, and 1.5 meters and reporting the mean depth error to a plane fitted to each measurement.
We find that FWL outperforms other modalities that do not use all the available degrees of freedom for modulating light (Table~\ref{tab:captured}).

\paragraph{Robustness to noise.} 
We evaluate our method using different per-pixel exposure settings (from 0.125 microseconds to 1 microsecond) while maintaining the same transmit power.
At shorter exposure settings the quality of depth estimation degrades (Figure~\ref{fig:exposure}); however, the proposed approach demonstrates better performance at low signal to noise ratios compared to generalized matched filtering.  

\subsection{Imaging Demonstrations}
\paragraph{Recovering depth and velocity.}
In Figure~\ref{fig:teaser} we show a scene illuminated by sunlight (through a window) containing static objects and a spinning hemisphere.
The radial velocity of the hemisphere is estimated to be a maximum of 25 m/s, which agrees with estimates produced using a high speed camera (see supplement).
Depth is recovered despite the presence of strong ambient light from the sun, which illuminates the scene through the window (see Figure~\ref{fig:results} (e)).

\paragraph{Challenging materials and room-sized scene.}
In Figure~\ref{fig:results} (a), we demonstrate reconstructing objects with subsurface scattering and detailed geometry and provide a comparison to generalized matched filtering using optical powers of 2 mW and 80 mW.
In each case the qualitative results in Figure~\ref{fig:results} show robust reconstruction of the scene geometry.
We see similar trends for a larger room-sized scene.

\paragraph{Imaging through translucent media.} 
In Figure~\ref{fig:results} (c), we show an example scene where a statuette is placed behind a translucent surface.
\parsarev{We recover both surfaces by estimating the time delays of the two Jones matrices with the greatest Frobenius norms.}

\paragraph{Strong ambient light.} 
We reconstruct the geometry of a tungsten halide light bulb while it is turned on (Figure~\ref{fig:results} (d)). 
While the bulb's spectrum includes the 1550 nm wavelength used for FWL, the system is extremely robust to ambient light due to the homodyne detection scheme.

%% file: sections/discussion.tex
\section{Discussion}\enlargethispage{6pt}
Coherent optical modems are a promising solution to make coherent lidar more accessible to researchers and practitioners.
Still, some barriers remain to widespread adoption of this technology.
For example, using fast optical modems (operating at tens to hundreds of GHz) requires domain knowledge to program and read out the transmitted and received waveforms. 
Currently, our hardware interface to the modem requires \parsafinal{$\sim$1} \parsa{second} to transfer data to a computer after each exposure; this limits the acquisition speeds of the current system to \parsa{more than a second} per scan point and thus limits the overall scan resolution. 
In future work we hope to address these obstacles and to demonstrate real-time capture of depth and velocity.
FWL may also offer benefits for other imaging scenarios, such as in the presence of scattering media, where sensitivity to motion, depth, and polarization could help to isolate unscattered light.

%% file: sections/acknowledgments.tex
\vspace*{-6pt}
\section*{\parsa{Acknowledgments}}

\parsa{We thank Ciena Corporation for loaning the optical modem and Praneeth Chakravarthula for helpful discussions. DBL and KNK acknowledge support from Ciena Corporation, and NSERC under the RGPIN, RTI, and Alliance programs. DBL also acknowledges support from the Canada Foundation for Innovation and the Ontario Research Fund. PM was supported by the Mitacs Accelerate program. MW was supported by an XSeed grant and TRANSFORM HF.}

\vspace*{-5pt}

%% file: sections/figs.tex
\begin{figure*}[t]
    \includegraphics[width=0.9\textwidth]{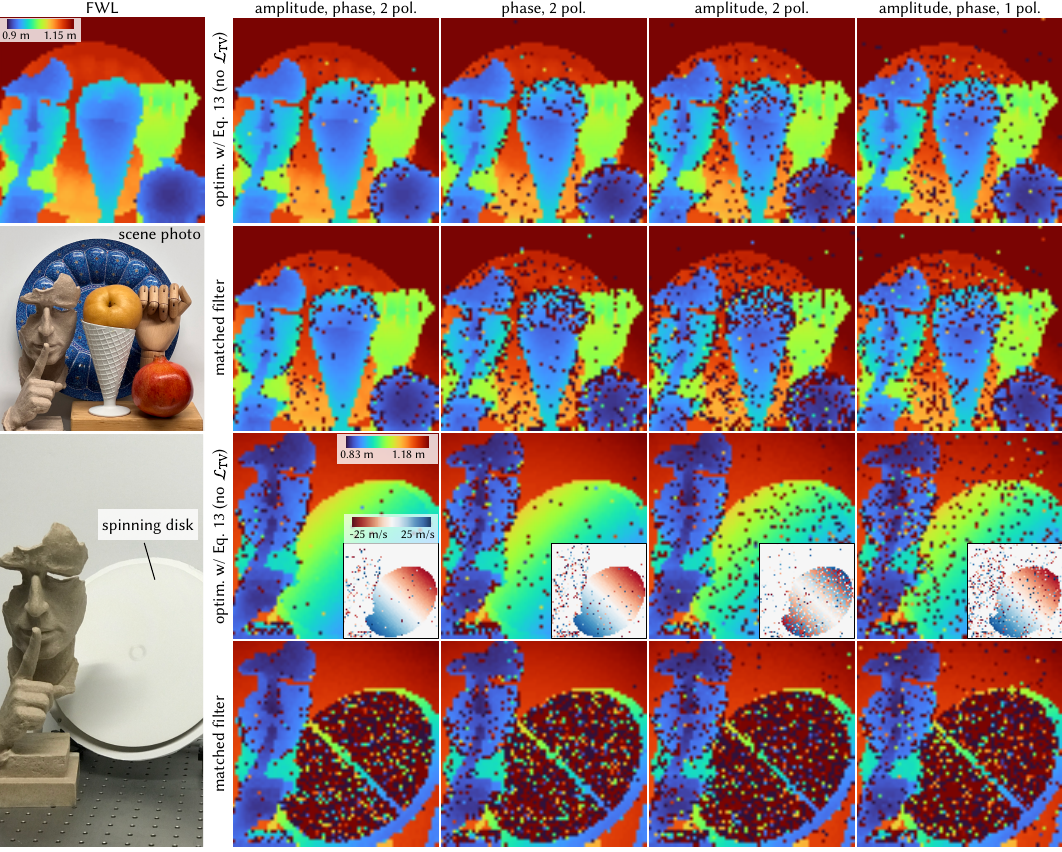}
    \vspace{-1em}
    \caption[]{Comparison between modulation schemes and optimization techniques. \textbf{(rows 1, 3)} We compare FWL (amplitude and phase modulation across two polarization channels) to phase-only modulation, amplitude-only modulation, and single-polarization modulation of phase and amplitude without $\mathcal{L}_\text{TV}$ (see supplement for implementation details). \textbf{(rows 2, 4)} Results using generalized matched filtering. FWL optimization based on Equation~\ref{eq:objective} shows more robust performance compared to other modulation schemes and the generalized matched filter, which also fails to recover accurate depth for Doppler-shifted pixels.}\label{fig:comparisons}
\end{figure*}

\begin{figure*}[t]
    \includegraphics[width=0.9\textwidth]{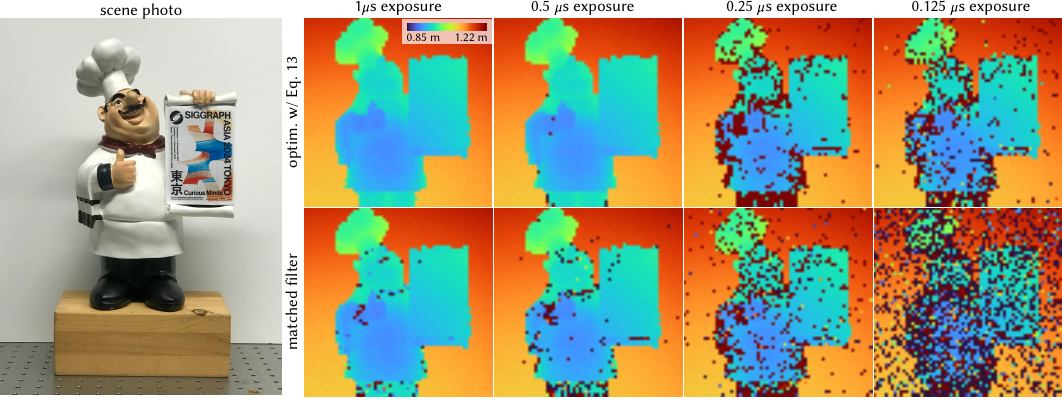}
\vspace{-1em}
 \caption[]{Optimization using Eq.~\ref{eq:objective} vs.\ generalized matched filtering for varying exposure times ranging from 1$\mu$s to 0.125$\mu$s. Even with exposures as short as 0.125$\mu$s the proposed technique recovers depth for most pixels while generalized matched filtering fails.}\label{fig:exposure}
\end{figure*}

\clearpage
 \input{sections/tables.tex}

\begin{figure*}[t]
    \includegraphics[width=0.96\textwidth]{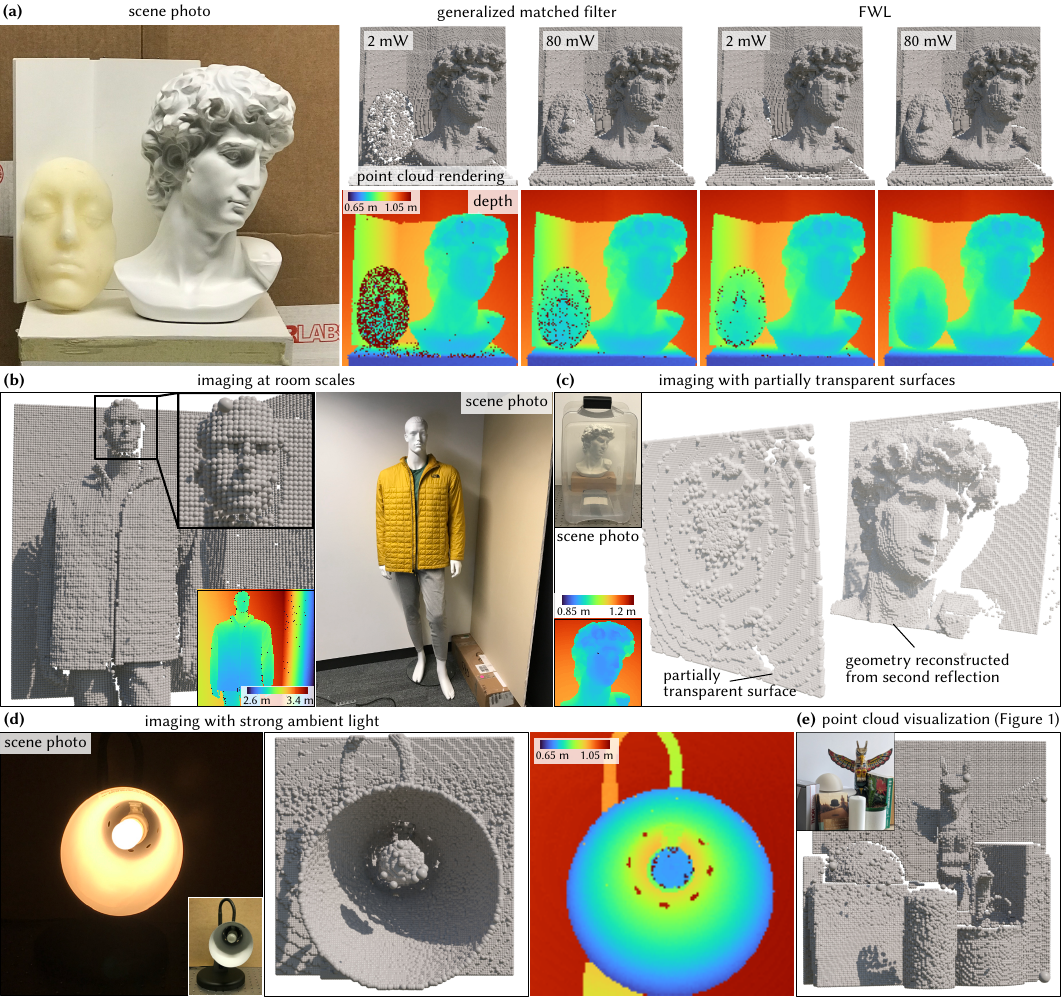}
    \vspace{-1em}
\caption[]{FWL in challenging imaging scenarios. \textbf{(a)} A diffuse statuette and mask made from translucent material are reconstructed with the laser power set to 2 mW and 80 mW. Using the proposed optimization (Equation~\ref{eq:objective}) with the 2 mW measurement recovers a qualitatively similar result to the generalized matched filtering scheme with 80 mW. \textbf{(b)} FWL captures a room-sized scene and recovers fine details on the jacket of the mannequin.
\textbf{(c)} A statuette is placed behind a translucent barrier and the reconstruction captures both the translucent surface as well as the 3D geometry of the statuette. 
\textbf{(d)} FWL scans a tungsten halide light bulb while it is turned on. Although the emission spectrum of the bulb includes the 1550 nm wavelength used by the system, FWL reconstructs both the lamp and the bulb itself. \textbf{(e)} A point cloud visualization of the scene in Figure~\ref{fig:teaser}. Captures for \textbf{(b)}, \textbf{(c)}, and \textbf{(d)} use a transmit power of 80 mW. }\label{fig:results}
\end{figure*}

\clearpage

%% file: sections/tables.tex
\begin{table*}[ht!]
    \caption{Evaluation of depth precision. We scan a planar surface and measure deviation of the measurements to a plane fitted to the surface. We compare to performance using the generalized matched filter; note that we omit the TV regularizer for this \parsa{evaluation} to assess per-pixel precision.}
    \vspace{-1em}
    \label{tab:captured}
    \centering
    \resizebox{0.9\textwidth}{!}{%
    \begin{tabular}{l|cc|cc|cc}
        \toprule  & \multicolumn{2}{c|}{mean depth error (mm)$\,\downarrow$} & \multicolumn{2}{c|}{\% of pixels with depth error < 2 mm$\,\uparrow$} & \multicolumn{2}{c}{\% of pixels with depth error < 6 mm$\,\uparrow$} \\ 
        \textbf{method} & joint estimation & gen.\ matched filter & joint estimation & gen.\ matched filter & joint estimation & gen.\ matched filter \\
        \midrule
        FWL & \textbf{4.43} & \textbf{9.93} & \textbf{65.20} & \textbf{64.77} &\textbf{98.50} & \textbf{97.78} \\
        dual-polarization phase & 9.31 & 24.99 & 57.16 & 55.40 & 97.91 & 94.92 \\
        dual-polarization and amplitude & 19.08 & 39.51 & 58.59 & 55.72 & 95.96 & 91.92 \\
        single-polarization phase and amplitude & 30.19 & 46.95 & 47.85 & 45.83 & 93.75 & 90.46\\
        \bottomrule
\end{tabular}}
\end{table*}

%% file: sections/supp.tex
\renewcommand\thesection{S\arabic{section}}
\renewcommand\thefigure{S\arabic{figure}}
\renewcommand\thetable{S\arabic{table}}
\renewcommand\theequation{S\arabic{equation}}

\section{Coherent Modulation and Demodulation}

\paragraph{Homodyne detection.}
For completeness, we provide a detailed mathematical description of the homodyne detection procedure that was introduced in Section 3 of the paper. 
Our derivation follows that of Ip et al.~\shortcite{ip2008coherent}.
However, for simplicity, we will ignore the effects of phase noise and amplified spontaneous emission noise in this derivation; these can be gathered into a single term consisting of complex Gaussian noise as we will note later.

Recall from the main text that the transmit and received electric fields can be written as 
\begin{align}
    \txfield(t) &= \sqrt{\txpower} e^{j \samplingfreq t} \, \mathbf{X}(t),\\
    \rxfield(t) &= \jonesmatrix \; \txfield(t-\timeshift) \; e^{j\freqshift t}\nonumber\\
                &= \sqrt{\txpower} \; \jonesmatrix \; \mathbf{X}(t-\timeshift) \; e^{j\freqshift t}e^{j \samplingfreq t},
\end{align}
where $\txpower$ is the transmit optical power.
Then, write the electric field corresponding to the local oscillator as
\begin{align}
    \lofield(t)[p] = \sqrt{\lopower} \; e^{j\samplingfreq t}, \quad p \in \{1, 2\},
\end{align}
where $p$ indexes the polarization channel and $P_\text{LO}$ is the transmit power of the local oscillator.

\begin{figure}[t]
\vspace{0pt}
\includegraphics{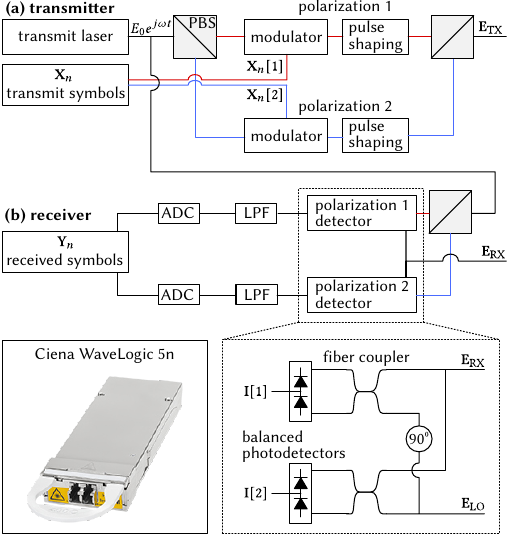}
\caption{Illustration of the transmit and receive path of a coherent optical modem. \textbf{(a)} The transmit signal comprises two polarization channels that are modulated using the transmit symbol sequence $\inputcodevector{n}$, pulse-shaped, and combined using a polarizing beam splitter (PBS). \textbf{(b)} The received signal is mixed with the local oscillator to perform \parsa{downconversion} and homodyne detection (inset, bottom right). Balanced photodetectors convert the intensity in each polarization channel to a photocurrent $\mathbf{I}[p]$ that is low-pass filtered (LPF) and sampled using an analog-to-digital converter (ADC) to capture the received symbol sequence $\outputcodevector{n}$. A photo of a coherent optical modem is shown in the bottom left.}\label{fig:modem}
\vspace{-3pt}

\end{figure}

As shown in Figure~\ref{fig:modem}, the received and local oscillator fields are combined and detected using two pairs of balanced photodiodes---one pair for each polarization channel. 
The photocurrent $\mathbf{I}[p]$ at the output of each balanced photodiode is given as 
\begin{align}
    \mathbf{I}[p] = \lvert \rxfield(t)[p] + \lofield(t)[p]\rvert^2 - \lvert \rxfield(t)[p] - \lofield(t)[p]\rvert^2,
    \label{eq:photocurrent}
\end{align}
where the sign change in the second term comes from a $90^{\circ}$ phase shifter and an additional $90^{\circ}$ phase shift induced by the fiber coupler (see Figure~\ref{fig:modem}).

Expanding the first term (and dropping dependencies on $t$ for convenience) yields 
\begin{align}
    & \lvert \rxfield[p] + \lofield[p]\rvert^2 \nonumber\\
    &= (\overline{\rxfield[p] + \lofield[p]})(\rxfield[p] + \lofield[p])\nonumber\\
    &= \txpower \,\lvert (\jonesmatrix\, \mathbf{X})[p]\rvert^2  + \lopower + (\overline{\rxfield}\lofield)[p] + (\rxfield\overline{\lofield})[p]\label{eq:firstterm}.
\end{align}
Expanding the second term in Equation~\ref{eq:photocurrent}, we similarly obtain
\begin{align}
    & \lvert \rxfield[p] - \lofield[p]\rvert^2 \nonumber\\
    &= \txpower \,\lvert (\jonesmatrix\, \mathbf{X})[p]\rvert^2  + \lopower - (\overline{\rxfield}\lofield)[p] - (\rxfield\overline{\lofield})[p]\label{eq:secondterm}.
\end{align}

Subtracting Equation~\ref{eq:secondterm} from Equation~\ref{eq:firstterm} gives
\begin{align}
    \mathbf{I}[p] &= 2(\overline{\rxfield}\lofield)[p] + 2(\rxfield\overline{\lofield})[p]\nonumber\\
         &= 2 \sqrt{\txpower}\sqrt{\lopower} \; \lvert (\jonesmatrix \, \mathbf{X})[p]\rvert \left(e^{-j(\freqshift t + \angle{(\jonesmatrix \, \mathbf{X})[p]})} + e^{j(\freqshift t + \angle{(\jonesmatrix \, \mathbf{X})[p])}} \right) \nonumber\\
         &= \underbrace{4 \sqrt{\txpower}\sqrt{\lopower} \; \lvert (\jonesmatrix \, \mathbf{X})[p]\rvert}_{\mathbf{A}[p]} \cos(\freqshift t + \underbrace{\angle{(\jonesmatrix \, \mathbf{X})[p]}}_{\boldsymbol{\phi}[p]}).
\end{align}
So the balanced photodetectors remove all steady-state signals. 
Any terms due to ambient light would also be canceled out in the balanced photodetection. 
The final photodetector current is a cosine function whose amplitude is the product of the transmit and receive amplitudes, the Jones matrix, and the transmitted symbol amplitudes. 
The frequency depends on the Doppler shift, and the phase of the signal depends on the phase of the Jones matrix entries and the transmitted symbols.

In practice, the coherent modem captures complex-valued samples of the photocurrent. 
That is, the photocurrent signal is passed through a splitter, and one signal copy is sampled directly while the other copy is delayed with a 90 degree phase shift and then sampled. 
The resulting signal is
\begin{align}
    \underbrace{\mathbf{A}[p]\cos(\freqshift t + \boldsymbol{\phi}[p])}_{\text{in-phase}} + \underbrace{\mathbf{A}[p]\sin(\freqshift t + \boldsymbol{\phi}[p])}_{\text{quadrature}}.
\end{align}
These two sampled signals are commonly called the in-phase and quadrature signal components, where the quadrature signal corresponds to the phase-delayed copy.

Finally, treating the in-phase and quadrature signals as the real and imaginary components of a complex-valued signal, respectively, yields
\begin{align}
    & \mathbf{A}[p]\cos(\freqshift t + \boldsymbol{\phi}[p]) + j\mathbf{A}[p]\sin(\freqshift t + \boldsymbol{\phi}[p]) \nonumber\\
    &= \mathbf{A}[p]e^{j(\freqshift t + \boldsymbol{\phi}[p])} \nonumber\\
    &\propto \rxfield[p] e^{-j\samplingfreq t}.
\end{align}
such that the homodyne detection procedure recovers a complex-valued signal that is proportional to the received electric field.
While we ignore noise in this derivation for simplicity, it is typically modeled using a complex Gaussian distribution. 
The dominating sources of noise are local oscillator shot noise and amplified spontaneous emission noise due to the Erbium-doped fiber amplifiers used to amplify the received laser light~\cite{ip2008coherent}.

\section{Supplemental Implementation Details}

\paragraph{Optimization.}
We implement the optimization using a two-stage procedure.
First, we note that in the absence of total variation regularization the objective function (Equation 13) can be minimized in a per-pixel fashion. 
For computational expediency, we conduct a first stage of optimization where $\jonesmatrix_{\discretetimeshift, \freqshift}$ are estimated for each pixel in parallel using only the sparsity regularizer.
We also assume the maximum plausible distance from the system to be 4 meters, and only optimize for the Jones matrices associated with the feasible time delays. 
We find that 50 iterations of optimization using Adam~\cite{kingma2015adam} with $\lambda_\text{sparse} = 10^{-1}$ for static scenes and $\lambda_\text{sparse} = 3\times 10^{-1}$ for dynamic scenes is sufficient for the estimated depth to converge. 
To avoid unnecessary computation during the optimization, we assume that $\jonesmatrix_{\discretetimeshift, \freqshift}=0$ for all $\freqshift \neq 0$ for scenes that are known to be static (i.e., no Doppler shift). In this case we optimize only the set of Jones matrices for which $\freqshift = 0$.
For static pixels, this optimization requires a few seconds per pixel using an NVIDIA A40 GPU. 
With our unoptimized implementation, processing each pixel with a Doppler shift requires roughly one minute on the same hardware.

In the second stage of optimization we add the total variation penalty; this procedure requires processing the entire image at once due to the dependencies between pixels. 
However, given the long sequence lengths of $\jonesmatrix_{\discretetimeshift, \freqshift}$ (typically several thousand samples along the temporal dimension), $\mathbf{E}_\text{TX}$ ($\approx 2^{16}$ symbols), plus the additional dimensions associated with the number of pixels and the entries of the Jones matrix, it is challenging to process the entire captured dataset at every iteration using full-batch gradient methods.
Instead, we stochastically sample a number of pixels and their neighbors to calculate each term of Equation 13, including the data term, the sparsity term, and the total variation penalty (we use a batch size of 1024 pixels at each iteration).
We find that this stage of the optimization converges within 500 iterations, for a total of 550 iterations of optimization including the first stage.
The second stage of optimization takes roughly 30 minutes to complete using an Nvidia A40 GPU.

Finally, we note that we do not apply the total variation penalty to dynamic scenes when estimating the Doppler shifts for the Jones matrices.
While we find the total variation penalty to be effective for the static scenes (e.g., Figure 9), the additional velocity dimension increases memory requirements, making it challenging to apply total variation to the large amount of data captured by the optical modem across all possible frequency shifts.
Finding efficient ways to work with the large quantities of data captured by an optical modem is an interesting direction for future work.

\paragraph{Implementation of other modulation schemes.}
We compare FWL to using phase-only modulation with two polarization channels, amplitude-only modulation with two polarization channels, and phase and amplitude modulation with one polarization channel.
The transmit power is kept the same as for FWL in all cases (including for the single polarization modulation scheme).

To implement the phase-only modulation scheme, we generate a symbol sequence with constant amplitude and uniformly distributed phases.
Since our intent is to emulate phase modulation-based lidars that are not sensitive to amplitude, we discard the amplitude information by normalizing the symbols at the receiver prior to estimating depth or velocity.
For amplitude-only modulation, we transmit symbols with constant phase and normally distributed amplitudes.
The reconstruction is performed by projecting the received symbols onto a complex-valued unit vector with same phase used for modulation. 
This procedure removes the phase information from the receiver.
Finally, for single-polarization phase and amplitude modulation, we simply discard one of the polarization channels at the receiver.

\begin{figure}[t]
    \vspace{5pt}
    \includegraphics[width=\columnwidth]{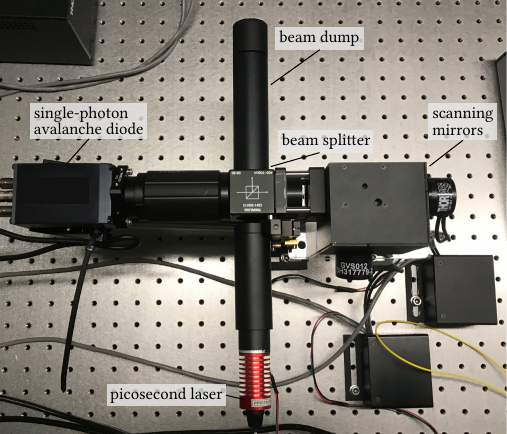}
    \caption{Prototype single-photon lidar system. A single-pixel, single-photon avalanche diode shares an optical path through a beamsplitter with a picosecond laser. A pair of scanning mirrors is used to raster scan the scene.}\label{fig:spad}
\end{figure}

\paragraph{Single-photon lidar system.} 
To compare FWL with single-photon lidar (Figure 2), we built a prototype single-photon lidar system, shown in Figure~\ref{fig:spad}. 
The prototype comprises a single-pixel single-photon avalanche diode (SPAD; MPD Fast-Gated Module), a beam splitter to separate the outgoing laser beam and incoming reflected light, a picosecond pulsed laser operating at 670 nm (Alphalas Picopower), a pair of scanning mirrors (Thorlabs GVS012), and a time-correlated single photon counter (TCSPS; PicoQuant PicoHarp 300). We set the laser pulse repetition rate to 10 MHz and configure the power to achieve approximately 500,000 photon counts per second, which is roughly the threshold at which non-linear pileup effects are still negligible~\cite{rapp2021high}.  

\section{Supplemental Results}

\begin{figure}[t]
    \includegraphics{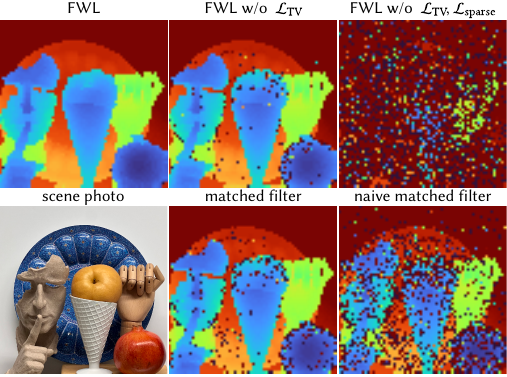}
    \caption{Performance of FWL regularization and matched filtering. FWL using both total variation and sparsity regularizers (top left) performs best for this scene containing a few static objects. The method without total variation (top middle) is more sensitive to speckle noise, and removing both regularizers results in poor performance. The generalized matched filter (bottom middle) performs better than ``naive'' matched filtering (bottom right), which uses cross correlations between the two corresponding polarization channels and fails to exploit cross-polarization information.}\label{fig:supp-static}
\end{figure}

\begin{figure*}
    \includegraphics{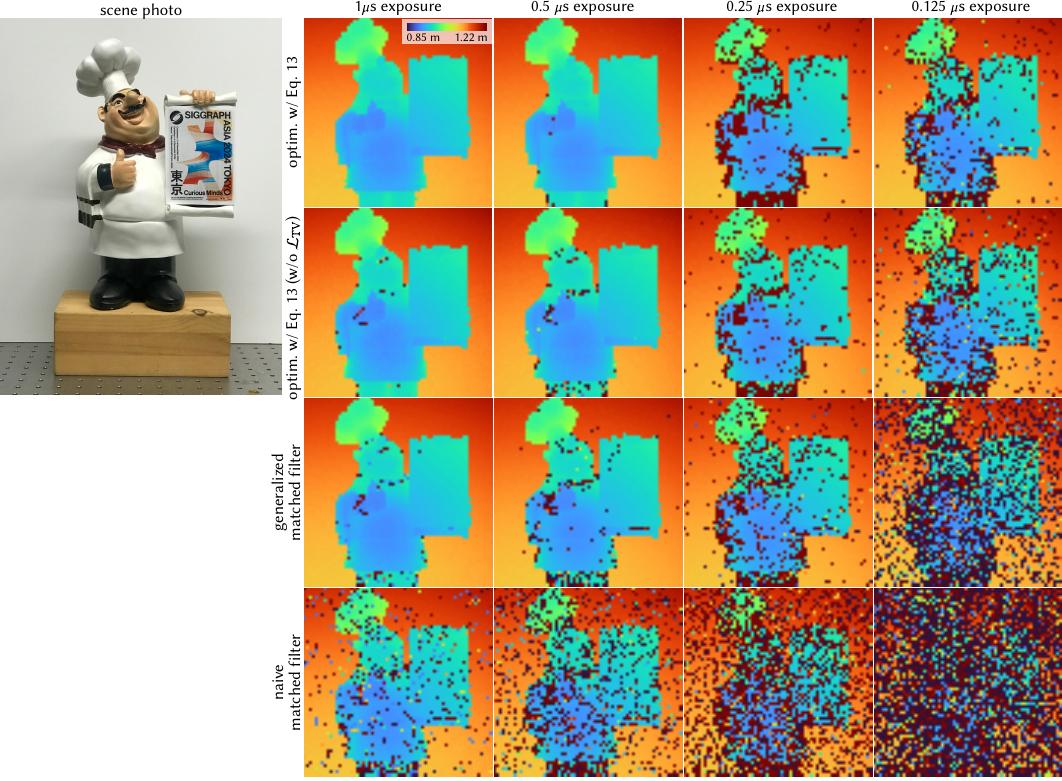}
    \caption{Additional comparisons of FWL and matched filtering vs.\ exposure time. We compare reconstruction using both sparsity and total variation regularization (row 1) to the optimization without total variation regularization (row 2).  We also compare to generalized matched filtering (row 3), which exploits cross polarization information, and naive matched filtering (row 4), which only correlates the two corresponding polarization channels. FWL using both regularizers produces depth maps with the fewest outliers as the exposure time decreases.}\label{fig:supp-chef}
\end{figure*}

\paragraph{Supplemental ablation studies.}
We provide additional results showing the performance of FWL with both total variation and sparsity regularization, without total variation regularization, and without any regularization (Figure~\ref{fig:supp-static}). We find that using both regularizers produces the best results for this static scene. 
We also compare to the performance of generalized matched filtering, which correlates the known transmit sequence at each polarization channel with each of the received polarization channels (i.e., exploits cross-channel polarization information). 
We find that this approach recovers depth estimates with fewer outliers than a naive form of matched filtering, which only correlates the corresponding transmit and receive polarization channels (also shown in Figure~\ref{fig:supp-static}).
We observe similar trends in Figure~\ref{fig:supp-chef}, which shows the same comparison for a range of exposure times. 
As the exposure time decreases, we find that FWL using both sparsity and total variation regularization produces depth maps with the fewest outliers compared to matched filtering or FWL without sparsity or total variation regularization.

\begin{figure*}
    \includegraphics[width=0.9\textwidth]{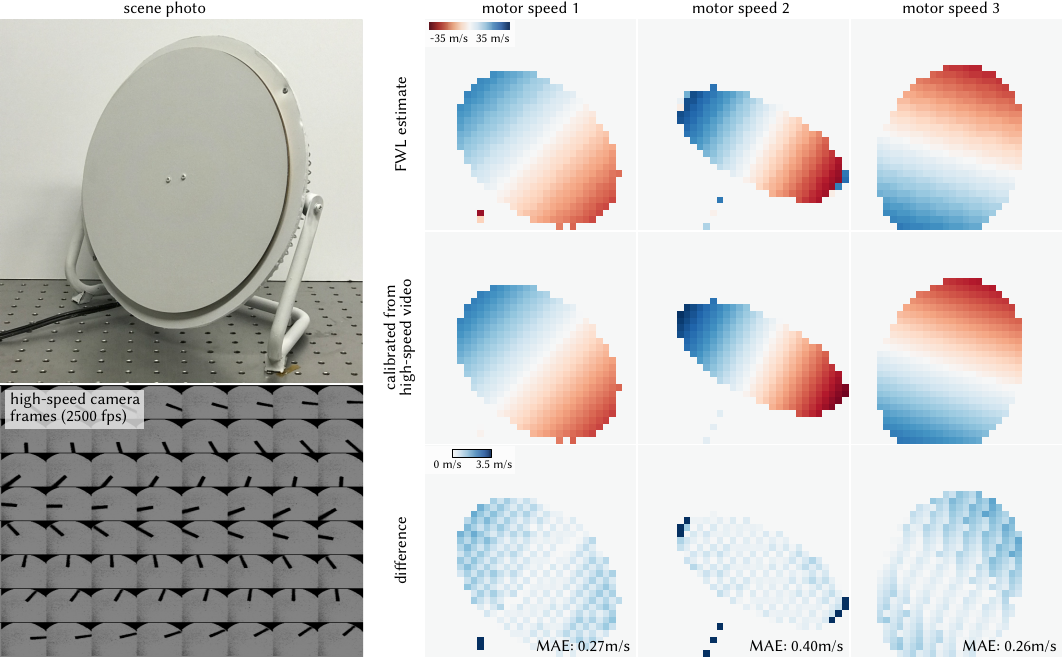}
    \caption{Assessment of the radial velocity predictions from the coherent optical modem for a spinning disk scene. We use a high-speed camera to capture frames of the spinning disk after adding a marker to track its position. Then, we fit a plane to the depth estimates from the coherent optical modem and use the rotational speed from the high-speed camera to estimate the radial velocity of the disk. We show the radial velocities computed in this fashion and using the Doppler shift from the optical modem for three different fan orientations and motor speeds. We find that the estimates from both methods agree to within a meter per second in terms of mean absolute error (MAE; right, bottom row). We use a transmit power of 10 mW for this experiment.}\label{fig:supp-doppler}
\end{figure*}

\paragraph{Quantitative evaluation of estimated radial velocity.}
To evaluate the estimated radial velocities shown in Figure 1 and Figure 9 of the main paper, we measure the rotational speed of the fan using a high speed camera. 
Then, we fit a plane to the scene, which contains a spinning disk that we retrofit to a fan motor. 
Using the plane fit, we compute the surface normal corresponding to each measured pixel, and we estimate the per-pixel radial velocities.
We find that this approach for estimating velocity agrees with our estimates using the Doppler shift to within a meter per second in terms of mean absolute error (MAE) as shown in Figure~\ref{fig:supp-doppler}.
Measurements from each approach are compared using the same spanning disk at different orientations and for different motor speed settings. 

\paragraph{Illustration of Figure 1 setup.}
We provide a labeled image depicting the capture setup used for Figure 1. In particular, we note that the scene was captured near a window and was illuminated by strong ambient light from the sun. 

\begin{figure*}
    \includegraphics[width=0.9\textwidth]{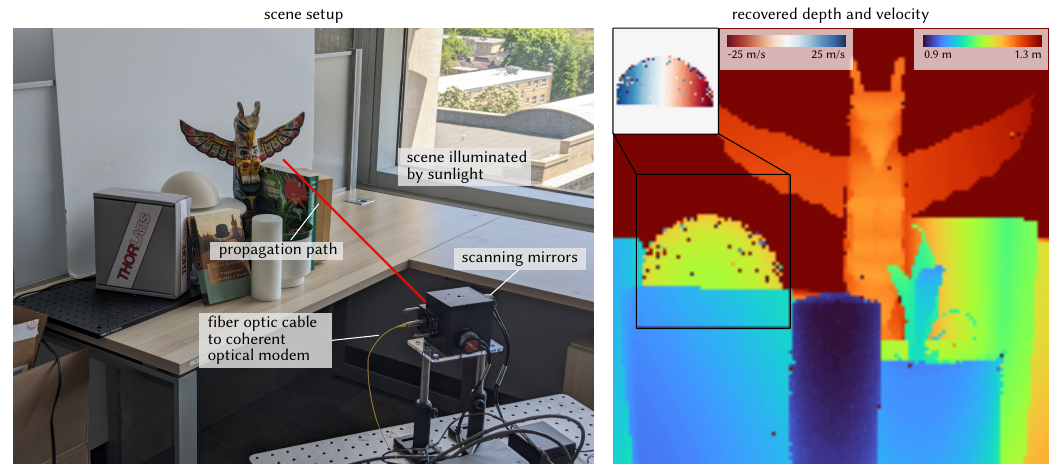}
    \caption{Capture setup of Figure 1. We show a labeled photo (left) depicting the capture setup for Figure 1, including illumination by sunlight through a window. The fiber optic cable to the modem, scanning mirrors, and propagation path are also labeled. For convenience, the depth and velocity reconstructions are reproduced from the main paper (right). }\label{fig:supp-teaser}
\end{figure*}